%% file: main.tex
\PassOptionsToPackage{table}{xcolor}

\documentclass[10pt,twocolumn,letterpaper]{article}

\usepackage[pagenumbers]{cvpr} 
\usepackage{tikz}
\usepackage{multicol}
\usetikzlibrary{positioning, shapes, arrows, backgrounds, fit, arrows.meta}
\usetikzlibrary{fit}
\usepackage[table]{xcolor} 
\usepackage{multirow}
\usepackage{float}
\usepackage{siunitx}
\usepackage{adjustbox}
\usepackage{longtable}
\usepackage{array}
\usepackage{makecell}


\definecolor{cvprblue}{rgb}{0.21,0.49,0.74}
\usepackage[pagebackref,breaklinks,colorlinks,allcolors=cvprblue]{hyperref}

\def\paperID{14000} 
\def\confName{CVPR}
\def\confYear{2025}

\title{On the robustness of multimodal language model towards distractions}

\author{
Ming Liu\thanks{pkulium@iastate.edu} \\
Iowa State University \\
Ames, Iowa \\
\and
Hao Chen\thanks{haoc3@andrew.cmu.edu} \\
Carnegie Mellon University \\
Pittsburgh, PA \\
\and
Jindong Wang\thanks{jwang80@wm.edu} \\
William \& Mary University \\
Williamsburg, VA \\
\and
Wensheng Zhang\thanks{wzhang@iastate.edu} \\
Iowa State University \\
Ames, IA
}

\begin{document}
\maketitle
\input{sec/0_abstract} 
\input{sec/1_intro}
\input{sec/2_related_work}
\input{sec/3_method}

\input{sec/4_experiment}
\input{sec/5_ablation}

\input{sec/6_discussion}
\input{sec/7_limitation_and_conclusion}
{
    \small
    \bibliographystyle{ieeenat_fullname}
    \bibliography{main}
}


\end{document}

%% file: sec/0_abstract.tex
\begin{abstract}

Although vision-language models (VLMs) have achieved significant success in various applications such as visual question answering, their resilience to prompt distractions remains as an under-explored area. Understanding how distractions affect VLMs is crucial for improving their real-world applicability, as inputs could have noisy and irrelevant information in many practical scenarios. This paper aims to assess the robustness of VLMs against both visual and textual distractions in the context of science question answering. Built on the \emph{ScienceQA} dataset, we developed a new benchmark that introduces distractions in both the visual and textual contexts. To evaluate the reasoning capacity of VLMs amidst these distractions, we analyzed the performance of ten state-of-the-art VLMs, including GPT-4o. Our findings reveal that most VLMs are vulnerable to various types of distractions, experiencing noticeable degradation in reasoning capabilities when confronted with distractions. Notably, models such as InternVL2 demonstrates a higher degree of robustness to these distractions. We also found that models exhibit greater sensitivity to textual distractions than visual ones. Additionally, we explored various mitigation strategies, such as prompt engineering, to counteract the impact of distractions. While these strategies improved model resilience, our analysis shows that there remain significant opportunities for improvement. 

\end{abstract}

%% file: sec/1_intro.tex
\section{Introduction}
\label{sec:introduction}

Despite the impressive capabilities of vision-language models (VLMs) in understanding images and generating human-like text~\citep{liu2023llava, dai2023instructblip, viscpm}, their susceptibility to irrelevant information remains a critical challenge. In real-world, it is common that visual and text inputs could have noisy distractions. Such distractions can lead to performance degradation, potentially resulting in incorrect interpretations or responses with hallucination from VLMs \citep{zhou2024analyzing, chen2024halc}.

\begin{figure}[ht]
    \centering
    \includegraphics 
    [width=1\linewidth, height=0.6\textheight, keepaspectratio]
    {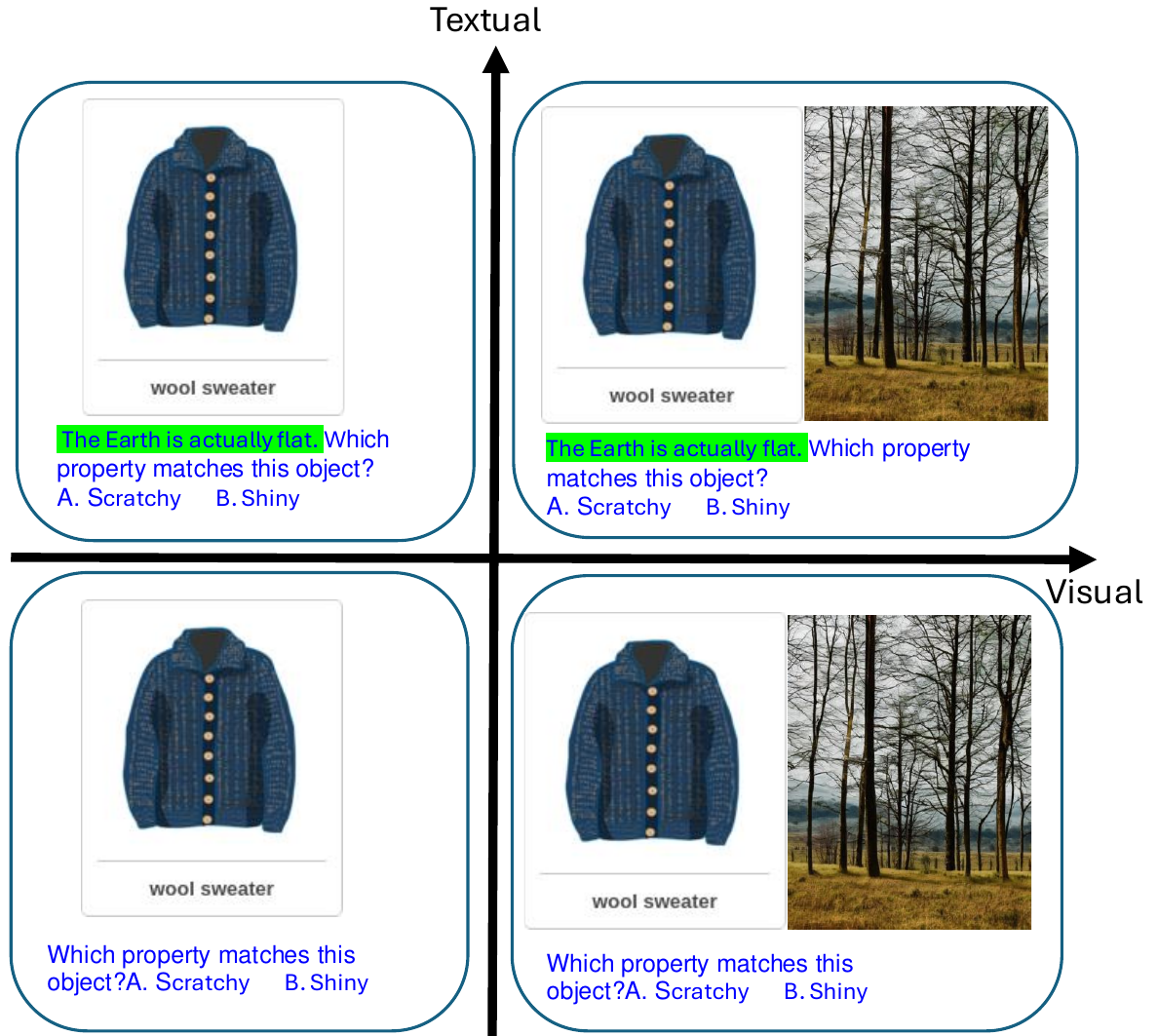}
    \vspace{-0.15in}
    \caption{Diagram illustrating various scenario of distraction we apply to the samples in \emph{Science-QA} dataset.}
    \label{fig:build dataset}
\vspace{-0.25in}
\end{figure}

Existing benchmarks for VLMs are typically designed under the assumption that inputs in both visual and textual domains are carefully curated without distractions. This assumption, however, fails to reflect real-world scenarios. Previous research~\citep{shi2023large} has demonstrated that large language models (LLMs) are vulnerable to textual distractions. With the rapid development of VLMs, it is crucial to understand how these models handle distractions not only in the textual domain but also in the visual domain. Compared to LLMs, VLMs face the additional challenge of potential distractions from \textit{bi-modal} inputs, making the situation more complex.

Moreover, current evaluation benchmarks~\citep{lu2022scienceqa,singh2019towards,lu2024mathvistaevaluatingmathematicalreasoning} do not adequately account for the presence of distractions within the input. They often emphasize clean and well-structured datasets, which do not mirror the complexities and noise inherent in real-world data. This oversight limits the ability to assess the true robustness and reliability of VLMs when deployed in practical settings where distractions are inevitable. Consequently, there is a pressing need for benchmarks that systematically introduce and evaluate various types of distractions to better understand and improve VLM performance under realistic conditions.

To address this gap, we present I-ScienceQA, a comprehensive benchmark designed to investigate the robustness of VLMs towards distractions. Our benchmark, built upon the ScienceQA dataset~\citep{lu2022scienceqa}, incorporates various types of distractions to simulate more realistic scenarios. Specifically, we aim to answer the following questions:

\begin{itemize}[leftmargin=2em]
\setlength\itemsep{0em}
     \item How vulnerable are VLMs towards distractions across different modalities? 
     \item Which modality, visual or textual, causes more degradation in model performance when distracted? 
     \item What techniques can mitigate the impact of distractions and improve the robustness of VLMs? 
\end{itemize}

To build \emph{I-ScienceQA}, we leveraged different generative models, including GPT-3.5-turbo~\citep{openai2024gpt35turbo} and Stable diffusion models~\citep{Rombach2021HighResolutionIS}. Our benchmark comprises 8,100 samples with four scenarios of distractions in both visual and textual domains. Specifically, we utilized stable diffusion models to generate visual distractions, such as neutral backgrounds, generic landscapes, abstract art, and everyday objects. For textual distractions, we employed GPT-3.5-turbo to produce textual distractions such as contradictory information, irrelevant details. This approach allowed us to simulate a wide range of real-world scenarios where VLMs might encounter noisy or irrelevant information. More information about the definition of distractions can be found in Appendix.

Through extensive evaluation of the various state-of-the-art VLMs, our key findings include:

\begin{itemize}[leftmargin=2em]
\setlength\itemsep{0em}
    \item VLMs exhibit varying degrees of vulnerability to distractions, with performance degradation observed across different models and scenarios (see Section~\ref{sec:experiment results}).
    \item Textual distractions tend to have a more significant impact on VLMs compared to visual distractions, particularly in the ``Add Hints'' scenario (see Section~\ref{sec:experiment results}).
    \item Larger models generally demonstrate better robustness against distractions, with some models like Internvl2 (8B) showing minimal performance drops in certain scenarios (see Section~\ref{subsec:model size}).
    \item Prompt engineering techniques or robust encoders offer limited enhancement to VLM performance against distractions, with their effectiveness varying across different models and tasks (see Section~\ref{subsec:denfending}). 
    \item The impact of bi-modal distractions (both visual and textual) on VLMs is nuanced, with some models showing consistent performance while others exhibiting minor fluctuations (see Section~\ref{subsec:bi modal distraction}).
\end{itemize}

Our research not only provides valuable insights into the current limitations of VLMs but also highlights potential areas for improvement in model design and training methodologies. By addressing these challenges, we can develop more robust and reliable VLMs for real-world applications.

%% file: sec/2_related_work.tex
\section{Related Work}
\label{sec:related_work}

\textbf{Model Evaluations} VLMs have traditionally been evaluated using standard Visual Question Answering (VQA) tasks such as TextVQA~\citep{singh2019towards}, VQAv2~\citep{vqav2}, and GQA~\citep{hudson2019gqa}. Mecently, studies like MM-Vet~\citep{yu2023mm}, POPE~\citep{li2023pope}, and MM-Bench~\citep{liu2023mmbench} have emerged to  evaluate VLMs, in key challenges such as hallucination, reasoning. These efforts have demonstrated that multimodal LLMs encounter significant issues, such as hallucination~\citep{guan2023hallusionbench} and insufficient robustness~\citep{fu2023mme}. In this papr, we introduce the I-ScienceQA benchmark, which highlights that even advanced VLMs, such as GPT-4o~\citep{gpt4v}, struggle with basic visual questions when irrelevant distractions are present in the input. 


\noindent\textbf{Robustness of VLM} To test the robustness of model reasoning, researchers have constructed arithmetic reasoning benchmarks by paraphrasing or rewriting sentences from clean datasets~\citep{patel2021nlp,kumar2021adversarial, shi2023large}.
Recent studies have increasingly concentrated on the adversarial robustness of VLMs~\cite{qi2024visual, carlini2024alignedneuralnetworksadversarially, schlarmann2023adversarial, zhao2023evaluating, dong2023robust}. \citet{schlarmann2023adversarial} demonstrate that imperceptible perturbations in input images can enable attackers to manipulate LVLMs into generating specific outputs. Visual adversarial attacks designed to jailbreak LVLMs are introduced in works such as~\citet{carlini2024alignedneuralnetworksadversarially} and~\citet{qi2024visual}. Recently, studies have focused on training adversary-robust vision encoders ~\cite{schlarmann2024robustclip, mao2023understanding}.

%% file: sec/3_method.tex
\section{Benchmark}

\subsection{Overview of \emph{I-ScienceQA}}


In order to create a comprehensive benchmark for assessing the robustness of VLMs, it is essential to introduce minor distractions while ensuring that the hints for solving the questions remain accessible in either the textual or visual context. Developing \emph{I-ScienceQA} presented several challenges. Firstly, ensuring the diversity and relevance of distractions across both visual and textual modalities required meticulous selection and generation strategies. Additionally, maintaining the semantic integrity of the original questions while injecting distractions demanded advanced techniques in data augmentation and validation. To overcome these challenges, we leveraged state-of-the-art generative models, such as GPT-3.5-turbo~\citep{openai2024gpt35turbo} for textual distractions and Stable diffusion models~\citep{Rombach2021HighResolutionIS} for visual distractions, ensuring that the introduced noise was both diverse and contextually appropriate. These efforts resulted in a robust and versatile benchmark that not only fills the gaps left by existing datasets but also provides a nuanced framework for evaluating and enhancing the resilience of VLMs in practical applications. In this paper, we introduce the I-ScienceQA benchmark, consisting of 8,100 samples distributed across four distraction scenarios.

\textbf{Data Collection}  \autoref{fig:build dataset} illustrates the models we utilized to construct the dataset. In our study, we employed  LLMs to introduce textual distractions and stable diffusion models to generate visual distractions. As depicted in \autoref{fig:build dataset}, we took use of GPT-3.5-turbo to generate short textual contexts or insert brief distractions into existing text. For the visual domain, we employed stable diffusion models to create various image distractions. We also applied masks to the main objects in existing images and added distractions to other areas to ensure that the models could still extract useful information to answer the questions. \autoref{fig:build dataset} shows the detailed process for data generation.

\textbf{Dataset Statistics} Built upon the ScienceQA dataset, our dataset is crafted as a comprehensive and diversified benchmark for evaluating the robustness of VLMs against distractions. In Appendix, we present samples from the dataset for some of the distraction types. There is the dataset statistics in Appendix. Specifically, the \emph{I-ScienceQA} dataset contains 8,100 samples, which include 4,000 text-based distractions and 4,100 image-based distractions. This dataset encompasses 4 scenarios of distractions. The data are collected from four types of sources including stable diffusion\citep{Rombach2021HighResolutionIS}, GPT-3.5, Unsplash API\citep{unsplashapi}, and PromeAI\citep{promeai2024}. It offers a broad spectrum of distractions. We believe that \emph{I-ScienceQA} can serve as a comprehensive benchmark for evaluating the robustness of VLMs. In the following sections, we will describe how we established the \emph{I-ScienceQA} benchmark.

\subsection{Data Collection and Augmentation Strategies}

\textbf{Scenario I: Add Image} After randomly selecting 2,000 samples from the test partition of examples in \emph{ScienceQA}~\citep{lu2022scienceqa} that originally do not include images, we added images to these samples to introduce visual contexts that test the model's ability to integrate and prioritize textual information when paired with unrelated visual content. We employed stable diffusion models to create these images. The types of images added are shown in Appendix and their definition can be found in Appendix. We generated a variety of images, ranging from neutral backgrounds to emotional contexts. In \autoref{fig:build dataset}, we present an example where original sample lacks image context, and it is then augmented with image generated from stable diffusion model. 

The selection of 2,000 samples was strategically chosen to facilitate an even distribution across eight subtypes of visual distractions under scenario of \textbf{Add Image}, allocating the same number of samples to each subtype(see Appendix). This approach ensures that each subtype of distraction is adequately represented, providing a balanced and comprehensive evaluation. Additionally, limiting the number of samples to 2,000 makes the dataset manageable in size, allowing for efficient processing and analysis. Random selection was employed to minimize selection bias and ensure that the distractions are uniformly distributed, enhancing the benchmark's reliability and validity. Similarly, for remaining scenarios, we adopted the same sample selection scheme. Each of those scenarios involved randomly selecting 2,000 samples from \emph{ScienceQA} and evenly distributing them across their respective distraction subtypes. 

\textbf{Scenario II: Insert Image}
After randomly selecting another 2,000 samples from the test partition of examples in \emph{ScienceQA}~\citep{lu2022scienceqa} that already include images, we inserted visual distractions to them to test the VLMs' robustness against visual noise and their ability to maintain focus on relevant elements. We mainly collected visual distraction images from the Unsplash API \citep{unsplashapi} and then combined them with the original images side by side. The types of images we collected are the same as in the previous section, as shown in Appendix. Additionally, we randomly selected 100 samples with large blank areas in the images from these 2,000 samples and employed diffusion model-based methods~\citep{promeai2024} to in-paint distractions into these blank areas. For this small subset, we considered inserting distractions such as flying objects or sitting pets. More details of this diffusion inpainting can be found in Appendix. In \autoref{fig:build dataset}, we show an example where there is existing visual context in the original sample, and an object is inserted by inpainting.

\textbf{Scenario III: Add Hint}
We also explored the integration of textual distractions. Inspired by the findings that large language models can be significantly distracted by irrelevant context~\citep{shi2023large}, we designed textual distractions using the GPT-3.5-turbo  to challenge the VLMs' ability to focus on relevant content. We first randomly selected 2,000 samples from the test partition of examples in \emph{ScienceQA}~\citep{lu2022scienceqa} that have the textual hint as ``N/A'' and then replace it with GPT-3.5-turbo generated content. In \autoref{fig:build dataset}, we present an example where there is no textual context as hints in the original sample, and it is augmented with textual hints generated from GPT-3.5-turbo. More details of this scenario of textual distraction can be found in Appendix.

\textbf{Scenario IV: Insert Hint}
We randomly selected 2,000 samples from the test partition of examples in \emph{ScienceQA}~\citep{lu2022scienceqa} where explicit textual hint is provided. Inserting distractions requires careful integration to challenge the models' capacity to maintain focus on the relevant information. These distractions are designed to test the model's resilience against misleading cues without completely diverging from the context. We employed the GPT-3.5-turbo to insert textual distractions. Unlike the previous section, we fed the existing textual hint from each sample to better leverage the LLMs' ability to create distractions based on the existing hint. In \autoref{fig:build dataset}, we present an example where there is existing textual hints in the original sample, and it is inserted with textual distractions generated from GPT-3.5-turbo. Types of distributions are elaborated in Appendix.

Each of these scenarios introduces a layer of complexity into the interaction between text and image, leveraging detailed contexts to test the model's ability to navigate and prioritize information effectively. Additionally, we ensured that all generated images and texts adhere to strict ethical guidelines, avoiding the inclusion of harmful, biased, or inappropriate content. By implementing rigorous filtering  and manual reviews, we maintain the integrity and responsibility of our benchmark, thereby preventing the introduction of unethical concerns.






%% file: sec/4_experiment.tex
\section{Experimental Setup}
\label{sec:experimental_setup}


\subsection{Models}


To evaluate the robustness of VLMs, we employ 14 advanced VLMs. Regarding model size, we consider both small and large models, with sizes ranging from 1B to 34B parameters. In terms of model accessibility, we include both open-source models such as LLaVA~\citep{liu2023llava} and proprietary models such as GPT-4o. VLMs include LLaVA-1.5 (7B, 13B)~\citep{liu2023llava}, InstructBLIP-Vicuna (7B, 13B)~\citep{dai2023instructblip}, Phi3-V (4B)~\citep{phi3vfinetuning2023}, InternVL2 (1B, 2B, 8B, 26B)~\citep{chen2024far}, CogVLM2 (19B)~\citep{hong2024cogvlm2visuallanguagemodels}, Qwen2-VL (2B, 8B)~\citep{wang2024qwen2vlenhancingvisionlanguagemodels}, and GPT-4o. 

\subsection{Evaluation Metrics}

To assess the robustness of the model $\mathcal{F}$, we utilize the following evaluation metrics:

\begin{itemize}[leftmargin=*, itemsep=0pt, topsep=0pt, parsep=2pt, partopsep=0pt]
    \item \textbf{Exact Match} The metric $\mathrm{Accuracy}(\mathcal{F}; \mathcal{D})$ represents the mean exact match score of the model $\mathcal{F}$ across all test instances $\mathcal{D}$, where $y_d$ is the correct output for instance $d$. The exact match score equally weights all individual test instances and is calculated as:
    \begin{align*}
        \mathrm{Accuracy}(\mathcal{F}; \mathcal{D}) = 
            \frac{
                \sum_{d \in \mathcal{D}} 
                    \mathbf{1}\left[
                        \mathcal{F}(d) = y_d
                    \right]
            }{
                |\mathcal{D}|
            }
    \end{align*}
    
    \item \textbf{Exact Match Degradation} This metric quantifies the impact of distractions on the model's performance. For an exact match score $A_{\mathcal{F}, \mathcal{D}}$ achieved by $\mathcal{F}$ on the original dataset $\mathcal{D}$ and its corresponding score $A_{\mathcal{F}, \mathcal{D}'}$ on the dataset with added distractions $\mathcal{D}'$, the degradation in performance is computed as:
    \begin{align*}
        \Delta \mathrm{Accuracy}(\mathcal{F}) = A_{\mathcal{F}, \mathcal{D}'} - A_{\mathcal{F}, \mathcal{D}},
    \end{align*}
        where $A_{\mathcal{F}, \mathcal{D}'}$ denotes the model's exact match score on samples with distractions. The value of $\Delta \mathrm{Accuracy}(\mathcal{F})$ is always less than or equal to zero ($\Delta \mathrm{Accuracy}(\mathcal{F}) \leq 0$). A value of zero indicates that the model's performance remains unchanged despite the introduction of distractions, showcasing high robustness. Negative values indicate a decline in performance caused by distractions, with lower values reflecting higher vulnerability to such distractions. Therefore, the closer $\Delta \mathrm{Accuracy}(\mathcal{F})$ is to zero, the more robust the model is against distractions.
\end{itemize}

\section{Experimental results}
 
\label{sec:experiment results}

\begin{table*}[ht]
  \centering
  \caption{Performance of various models under different scenarios of distractions. The \textit{Original} columns display the exact match scores on the samples of the \emph{ScienceQA} benchmark. The \textit{Distraction} columns present corresponding results on the \emph{I-ScienceQA} benchmark, including both exact match scores and exact match degradation (shown in parentheses). Values marked as \textit{N/A} indicate that the model requires both text and image inputs and are therefore excluded from evaluation under that section.}
  \label{tab:overall results}
  \resizebox{\textwidth}{!}{
  \begin{tabular}{l|cc|cc|cc|cc}
    \toprule
    \textbf{Model} & \multicolumn{2}{c|}{\textbf{Add Image(\%)}} & \multicolumn{2}{c|}{\textbf{Insert Image(\%)}} & \multicolumn{2}{c|}{\textbf{Add Hints(\%)}} & \multicolumn{2}{c}{\textbf{Insert Hint(\%)}} \\
    \cline{2-9}
    & \textbf{Original} & \textbf{Distraction} & \textbf{Original} & \textbf{Distraction} & \textbf{Original} & \textbf{Distraction} & \textbf{Original} & \textbf{Distraction} \\
    \midrule
    Phi3v (4B) & N/A & 91.15 & N/A & 83.52 & N/A & N/A & N/A & N/A \\
    Instructblip (7B) & N/A & 41.05 & N/A & 35.45 & N/A & N/A & N/A & N/A \\
    Instructblip (13B) & N/A & 47.26 & N/A & 47.80 & N/A & N/A & N/A & N/A \\
    Qwen2-VL-Instruct (2B) & 63.30 & 63.30 (0.00) & 63.80 & 63.26 (-0.54) & 60.80 & 54.45 (-6.35) & 72.45 & 64.20 (-8.25) \\
    Qwen2-VL-Instruct (7B) & 83.10 & 83.10 (0.00) & 68.40 & 68.08 (-0.32) & 75.65 & 68.00 (-7.65) & 77.40 & 74.10 (-3.30) \\
    Llava (7B) & 71.30 & 68.05 (-3.25) & 68.75 & 66.36 (-2.39) & 69.70 & 63.80 (-5.90) & 70.55 & 69.30 (-1.25) \\
    Llava (13B) & 72.90 & 72.00 (-0.90) & 72.10 & 69.60 (-2.50) & 72.15 & 67.45 (-4.70) & 73.10 & 71.80 (-1.30) \\
    Llava (34B) & 88.05 & 87.50 (-0.55) & 81.55 & 79.51 (-2.04) & 84.65 & 82.65 (-2.00) & 85.50 & 83.00 (-2.50) \\
    Internvl2 (1B) & 85.60 & 79.70 (-5.90) & 88.10 & 83.47 (-4.63) & 87.80 & 80.55 (-7.25) & 85.90 & 82.85 (-3.05) \\
    Internvl2 (2B) & 91.35 & 86.75 (-4.60) & 93.50 & 90.23 (-3.27) & 91.40 & 82.35 (-9.05) & 93.65 & 91.50 (-2.15) \\
    Internvl2 (8B) & 95.45 & \textbf{94.45} (-1.00) & 96.90 & 94.23 (-2.67) & 94.80 & \textbf{93.60} (-1.20) & 97.60 & 95.90 (-1.70) \\
    Internvl2 (26B) & 95.35 & 93.40 (-1.95) & 97.40 & 95.14 (-2.26) & 95.20 & 92.80 (-2.40) & 97.55 & 96.55 (-1.00) \\
    CogVLM2 (19B) & 73.30 & 71.70 (-1.60) & 89.35 & 87.47 (-1.88) & 78.60 & 70.50 (-8.10) & 84.15 & 80.85 (-3.30) \\
    GPT-4o & 93.50 & 93.00 (-0.50) & 80.70 & 78.56 (-2.14) & 89.50 & 87.50 (-2.00) & 86.00 & 84.05 (-1.95) \\
    \bottomrule
  \end{tabular}
  }
\end{table*}

\autoref{tab:overall results} presents a comprehensive evaluation of various  models across different scenarios of distractions, measured under both original and distraction settings. For each scenario, the exact match degradation due to distractions is quantified in parentheses, providing insight into the robustness of each model against distractions. \textbf{The results exhibit differing degrees of degradation in model performance when exposed to distractions, highlighting the variability in the models' abilities to focus on relevant data}. It is important to note that the Phi3v and InstructBLIP models, which can only process inputs containing both textual and visual components, were evaluated exclusively on the \textbf{Add Image} and \textbf{Insert Image} scenarios. In this discussion, we analyze the models' performances in each scenario of distraction, focusing on both the exact match score and the exact match degradation score.

Firstly, in the \textbf{Add Image} scenario, models are evaluated on their ability to handle additional visual distraction. Notably, the Internvl2 (8B) model achieves the highest performance in the distraction scenario with a score of 94.45, exhibiting a minimal decrease of $-1.00$ from its original score of 95.45. Similarly, GPT-4o maintains high performance with an exact match score of 93.00 and a slight reduction of $-0.50$. In contrast, smaller models like Llava (7B) and Internvl2 (1B) show more significant drops in performance, with exact match scores of 68.05 ($-3.25$) and 79.70 ($-5.90$), respectively. These results suggest that \textbf{larger models tend to be more robust against visual distractions in this context}.

In the \textbf{Insert Image} scenario, where visual distraction are embedded into existing visual input, the performance trends are consistent. The Internvl2 (8B) model again demonstrates robustness with a exact match scores of 94.23 and a decrease of $-2.67$ from its original score of 96.90. Interestingly, the Qwen2-VL-Instruct (2B) and Qwen2-VL-Instruct (7B) models show minimal performance degradation, with exact match scores of 63.26 ($-0.54$) and 68.08 ($-0.32$), respectively. Despite smaller reduction in exact match score, these models have worse performance. 

When examining the \textbf{Add Hint} scenario, which involves injected textual distraction, the impact of distractions becomes more pronounced. Most models experience larger decreases in performance. The Internvl2 (2B) model, for instance, has an exact match score of 82.35, reflecting a significant drop of $-9.05$ from its original score of 91.40. Even the higher-performing Internvl2 (8B) and Internvl2 (26B) models face reductions to exact match scores of 93.60 ($-1.20$) and 92.80 ($-2.40$), respectively. These findings highlight that \textbf{adding textual distractions poses a greater challenge to the models compared to visual distractions, possibly due to the complexity of processing  textual information}. 

Lastly, in the \textbf{Insert Hint} scenario, where textual distractions are interspersed within existing text, models generally show moderate performance degradation. The Internvl2 (8B) model maintains a high exact match score of 95.90, with a decrease of $-1.70$ from its original score of 97.60. Similarly, GPT-4o achieves a exact match scores of 84.05, reflecting a decrease of $-1.95$. However, models like Qwen2-VL-Instruct (2B) exhibit a larger drop to a exact match scores of 64.20 ($-8.25$), indicating vulnerability to inserted textual distractions. These results suggest that \textbf{while some models are adept at managing inserted hints, others may struggle, potentially due to differences in their attention mechanisms or the diversity of their training data}.

%% file: sec/5_ablation.tex
\vspace{-0.1in}
\section{Experimental analysis}
\label{sec:experiment analysis}
\subsection{Model Size}

\begin{figure}[ht]
  \centering
  \begin{subfigure}{0.48\textwidth}
    \centering
    \includegraphics[width=\textwidth]{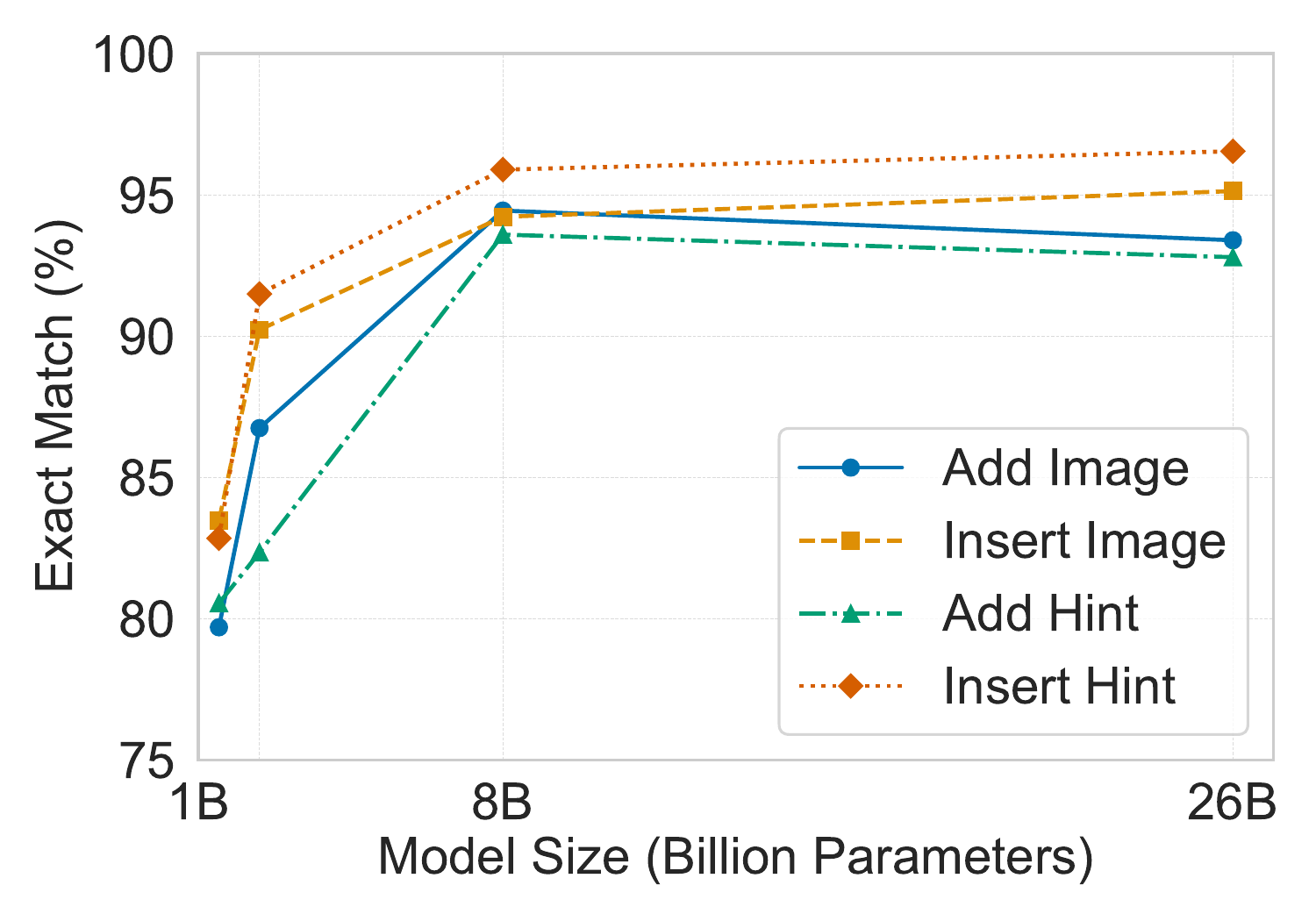}
    \label{fig:internvl2_performance_line}
  \end{subfigure}
  \hfill
  \begin{subfigure}{0.48\textwidth}
    \centering
    \includegraphics[width=\textwidth]{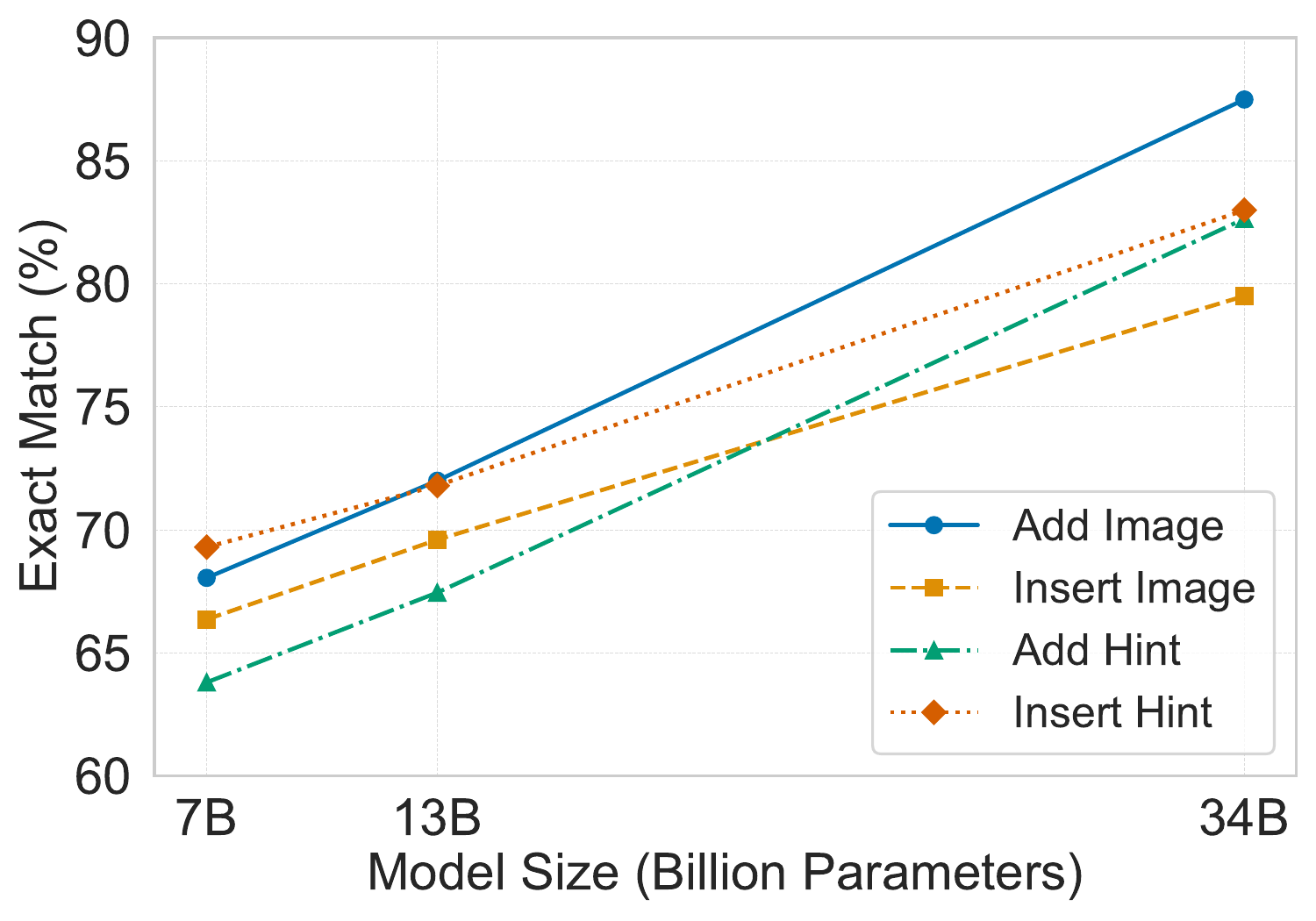}
    \label{fig:llava_performance_line}
  \end{subfigure}
  \vspace{-0.15in}
  \caption{Comparison of Exact Match Score for Internvl2(top) and Llava Models(bottom).}
  \label{fig:model size}
\vspace{-0.15in}
\end{figure}

Observing the performance across different model sizes in \autoref{fig:model size}, we notice that as the model size increases, there is a general improvement in performance across all scenarios. For the Internvl2 models, we consider four different sizes: 1B, 2B, 8B, and 26B parameters. In the \textbf{Add Image} scenario, the exact match scores increase from 79.70 for the 1B model to 94.45 for the 8B model, with a slight decrease to 93.40 at the 26B model. In the \textbf{Insert Image} scenario, performance improves steadily from 83.47 (1B) to 95.14 (26B). For the \textbf{Add Hints} scenario, scores rise from 80.55 (1B) to 93.60 (8B), then slightly decrease to 92.80 (26B). In the \textbf{Insert Hint} scenario, scores increase from 82.85 (1B) to 96.55 (26B). These results indicate that increasing the model size generally enhances performance, particularly up to the 8B parameter. The slight decrease or plateau in performance at the 26B size for some scenarios suggests that beyond a certain point, increasing model size yields diminishing performance or requires more sophisticated training techniques to leverage the additional parameters effectively.
\label{subsec:model size}

Similarly, for the Llava models with sizes 7B, 13B, and 34B, we observe performance trends in the distraction scenarios that reflect improvement with increased model size. In the \textbf{Add Image} scenario, scores increase from 68.05 (7B) to 87.50 (34B). In the \textbf{Insert Image} scenario, performance improves from 66.36 (7B) to 79.51 (34B). For the \textbf{Add Hint} scenario, scores rise from 63.80 (7B) to 82.65 (34B). In the \textbf{Insert Hint} scenario, scores increase from 69.30 (7B) to 83.00 (34B). The Llava models also show a clear trend of performance improvement with increased model size across all scenarios. The performance gains are more pronounced between the 7B and 34B models, suggesting that larger models can better handle distractions and integrate additional information effectively.

Comparing both models, the Internvl2 models generally outperform the Llava models at similar parameter sizes, especially in higher model sizes. For instance, the Internvl2 (8B) model achieves higher distraction scores than the Llava (13B) model across all scenarios, indicating that the Internvl2 architecture or training data may be more efficient in leveraging parameters for scenario performance. These observations underscore the significance of model scaling in enhancing performance, but they also highlight that architecture design and training data are crucial in maximizing the benefits of increased model size. 

\subsection{Analysis on training dataset 
 and model architecture}
\label{subsec:training data and model architecture}
 
\input{figure/dataset_model_encoder}



\subsection{Defending against distractions}
 \label{subsec:denfending}
 
\begin{table*}[ht]
  \centering
  \small \resizebox{\textwidth}{!}{
  \begin{tabular}{l|cc|cc|cc|cc}
    \toprule
    \textbf{Model} & \multicolumn{2}{c|}{\textbf{Add Image (\%)}} & \multicolumn{2}{c|}{\textbf{Insert Image (\%)}} & \multicolumn{2}{c|}{\textbf{Add Hints (\%)}} & \multicolumn{2}{c}{\textbf{Insert Hint (\%)}} \\
    \cmidrule(lr){2-3} \cmidrule(lr){4-5} \cmidrule(lr){6-7} \cmidrule(lr){8-9}
    & No Defense & Defense & No Defense & Defense & No Defense & Defense & No Defense & Defense \\
    \midrule
    Qwen2-VL-Instruct (2B) & 63.30 & 73.80 & 63.26 & 65.60 & 54.45 & 62.60 & 64.20 & 70.35 \\
    Qwen2-VL-Instruct (7B) & 83.10 & 81.35 & 68.08 & 68.60 & 68.00 & 68.05 & 74.10 & 74.90 \\
    CogVLM2 (19B)           & 71.70 & 70.15 & 87.47 & 85.18 & 70.50 & 70.20 & 80.85 & 79.10 \\
    \bottomrule
  \end{tabular}
  }
\caption{Exact match scores achieved by the models using a naive prompt compared to a prompt with instructions to ignore distractions.}
\label{tab:prompt defense}
\end{table*}

The findings in \autoref{tab:prompt defense} demonstrate that although prompt engineering techniques—such as adding instructions to guide the model's focus toward the question and away from distractions—can partially mitigate the effects of distractions, models still struggle to ignore them. For instance, in the \textbf{Add Image} scenario, the performance of Qwen2-VL-Instruct (2B) improves from 63.30 to 73.80 when defense mechanisms are applied, indicating that appropriate prompts can enhance the model's focus on relevant information. Similarly, in the \textbf{Insert Hint} scenario, the same model's performance increases from 64.20 to 70.35 with defense strategies.

However, the improvements are not uniform across all models and tasks. The Qwen2-VL-Instruct (7B) model shows a slight decrease in performance in the \textbf{Add Image} scenario when defenses are applied, dropping from 83.10 to 81.35. This suggests that the effectiveness of defense mechanisms may vary depending on the model's architecture and size. Moreover, the CogVLM2 (19B) model exhibits a minor reduction in performance across most tasks with defense prompts, indicating that larger models are not necessarily better at ignoring distractions when prompted to do so. For example, in the \textbf{Insert Image} scenario, its performance decreases from 87.47 to 85.18 even with defense.

\begin{table*}[ht]
  \centering
  \small
  \caption{Exact Math Scores achieved by LLava Models with different vision encoders.}
\label{tab:robust vision encoder}
  \resizebox{\textwidth}{!}{
  \begin{tabular}{l|cc|cc|cc|cc}
    \toprule
    \textbf{Model (Vision Encoder)} & \multicolumn{2}{c|}{\textbf{Add Image (\%)}} & \multicolumn{2}{c|}{\textbf{Insert Image (\%)}} & \multicolumn{2}{c|}{\textbf{Add Hints (\%)}} & \multicolumn{2}{c}{\textbf{Insert Hint (\%)}} \\
    \cmidrule(lr){2-3} \cmidrule(lr){4-5} \cmidrule(lr){6-7} \cmidrule(lr){8-9}
    & Original & Distraction & Original & Distraction & Original & Distraction & Original & Distraction \\
    \midrule
    LLava-7B (Robust-clip) & N/A & 70.40 & 67.30 & 63.55 & 71.78 & 67.48 & 64.31 & 63.25 \\
    LLava-7B (Clip)        & N/A & 69.55 & 68.95 & 64.32 & 76.22 & 72.49 & 64.39 & 63.25 \\
    \bottomrule
  \end{tabular}
  }
\end{table*}

Table~\ref{tab:robust vision encoder} summarizes the performance of LLava-7B models with two different vision encoders: \textit{robust-clip}~\citep{schlarmann2024robustclip} and \textit{clip}~\citep{ilharco_gabriel_2021_5143773}. The models are evaluated across original and distraction scenarios, focusing on the effects of adding or inserting images and hints. Since the LLava-7B model with \textit{robust-clip} can only process inputs that include both text and visual content, samples without images were excluded from this evaluation. The \textit{robust-clip} encoder only outperforms the \textit{clip} encoder slightly in the \textbf{Add Image} scenario by 0.85. In other scenario, the performance of the \textit{robust-clip} encoder is lower than that of the \textit{clip} encoder. These findings suggest that \textit{robust-clip} shows very limited efficacy in defending against visual distractions.

These results suggest potential areas for future improvements in model training and design. Developing more effective prompting techniques and enhancing model architectures could help models better filter out irrelevant information. Additionally, incorporating training data that specifically addresses the handling of distractions may improve models' robustness in real-world applications where irrelevant data is commonplace.

\subsection{Bi-Modal Distraction}
\label{subsec:bi modal distraction}

\begin{table}[ht]
  \centering
  \small
  \caption{Exact Match Score under bi-modal distractions.}
  \label{tab:bi_modal_injection}
  \begin{tabular}{lcccc}
    \toprule
    \multirow{2}{*}{\textbf{Model}} & \multicolumn{2}{c}{\textbf{Add Hint}} & \multicolumn{2}{c}{\textbf{Insert Hint}} \\
    \cmidrule(lr){2-3} \cmidrule(lr){4-5}
    & No Img & With Img & No Img & With Img \\
    \midrule
    Qwen2-VL-2B & 57.68 & 57.68 & 72.86 & 72.86 \\
    Qwen2-VL-7B & 71.43 & 71.43 & 88.57 & 88.57 \\
    CogVLM2-19B & 57.45 & 56.91 & 78.31 & 79.35 \\
    \bottomrule
  \end{tabular}
\end{table}
The results in~\autoref{tab:bi_modal_injection} examine the models' performances under conditions where textual distractions are present, with and without the simultaneous presence of visual distractions. Specifically, the ``No Image'' columns represent scenarios with only textual distractions, while the ``With Image'' columns include both textual and visual distractions. 

Analyzing the data, we observe that the performance of Qwen2-VL-Instruct (2B) and Qwen2-VL-Instruct (7B) remains unchanged between the ``No Image'' and ``With Image'' conditions across both ``Add Hint'' and ``Insert Hint'' scenarios. This suggests that the addition of visual distractions does not significantly impact these models when textual distractions are already present. In contrast, the CogVLM2 (19B) model shows a slight decrease in performance from 57.45\% to 56.91\% in the ``Add Hint'' scenario when an image is added, indicating a minor negative effect of visual distractions in conjunction with textual ones. Interestingly, in the ``Insert Hint'' scenario, its performance slightly improves from 78.31\% to 79.35\% with the addition of an image, suggesting that under certain conditions, visual distraction might compete with textual distraction.

Overall, these findings imply that the models' abilities to handle bi-modal distractions are nuanced. While some models maintain consistent performance regardless of the presence of additional visual information, others exhibit minor fluctuations. This highlights the importance of designing models that can effectively integrate and prioritize multi-modal inputs, ensuring robustness in environments where distractions are prevalent across different modalities.

%% file: figure/dataset_model_encoder.tex
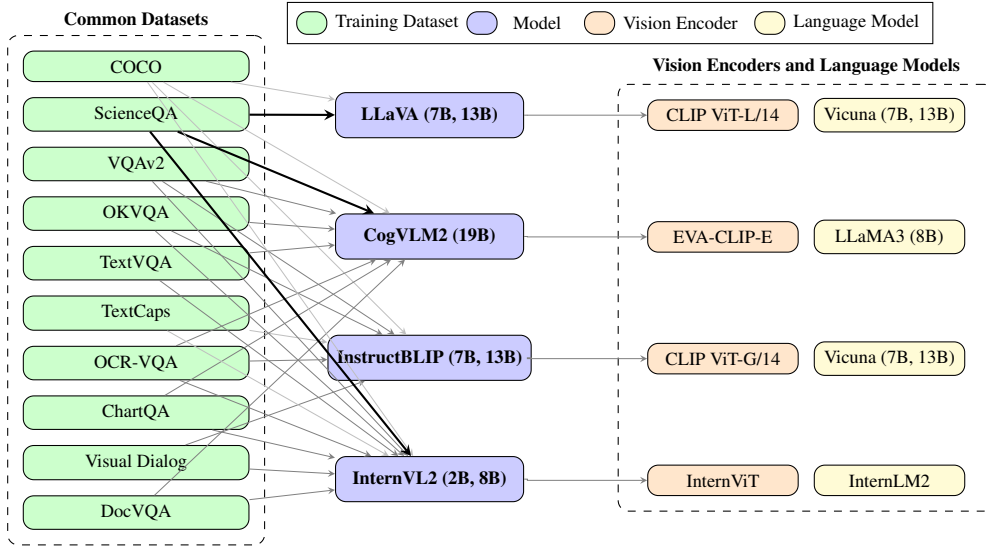
\begin{figure*}[ht]
\centering
\begin{tikzpicture}[
    scale=0.65, 
    node distance=0.3cm and 1.2cm, 
    dataset/.style={
        rectangle,
        draw,
        rounded corners,
        align=center,
        minimum height=0.4cm, 
        minimum width=3cm,
        font=\scriptsize 
    },
    pretrain/.style={
        dataset,
        fill=green!20
    },
    model/.style={
        rectangle,
        draw,
        rounded corners,
        align=center,
        minimum height=0.6cm, 
        minimum width=2.5cm,
        fill=blue!20,
        font=\bfseries\scriptsize 
    },
    vision/.style={
        rectangle,
        draw,
        rounded corners,
        align=center,
        minimum height=0.4cm, 
        minimum width=2cm,
        fill=orange!20,
        font=\scriptsize 
    },
    language/.style={
        rectangle,
        draw,
        rounded corners,
        align=center,
        minimum height=0.4cm, 
        minimum width=2cm,
        fill=yellow!20,
        font=\scriptsize 
    },
    arrow/.style={
        ->,
        very thin,
        >=stealth,
        draw=gray
    },
    arrow_light/.style={
        ->,
        very thin,
        >=stealth,
        draw=gray!50
    },
    highlight/.style={
        ->,
        black,
        thick,
        >=stealth
    },
    title/.style={
        font=\bfseries\small
    },
    legend/.style={
        rectangle,
        draw,
        align=left,
        fill=white,
        font=\scriptsize 
    }
]

\node[model] (LLaVA) at (0,3) {LLaVA (7B, 13B)};
\node[model, below=1cm of LLaVA] (CogVLM2) {CogVLM2 (19B)};
\node[model, below=1cm of CogVLM2] (InstructBLIP) {InstructBLIP (7B, 13B)};
\node[model, below=1cm of InstructBLIP] (InternVL2) {InternVL2 (2B, 8B)};

\node[pretrain] (coco) at (-6,4) {COCO};
\node[pretrain, below=0.2cm of coco] (scienceqa) {ScienceQA};
\node[pretrain, below=0.2cm of scienceqa] (vqav2) {VQAv2};
\node[pretrain, below=0.2cm of vqav2] (okvqa) {OKVQA};
\node[pretrain, below=0.2cm of okvqa] (textvqa) {TextVQA};
\node[pretrain, below=0.2cm of textvqa] (textcaps) {TextCaps};
\node[pretrain, below=0.2cm of textcaps] (ocrvqa) {OCR-VQA};
\node[pretrain, below=0.2cm of ocrvqa] (chartqa) {ChartQA};
\node[pretrain, below=0.2cm of chartqa] (visualdialog) {Visual Dialog};
\node[pretrain, below=0.2cm of visualdialog] (docvqa) {DocVQA};

\node[vision] (LLaVA_vision) at (6,3) {CLIP ViT-L/14};
\node[language, right=0.2cm of LLaVA_vision] (LLaVA_lang) {Vicuna (7B, 13B)};

\node[vision, below=1.2cm of LLaVA_vision] (CogVLM2_vision) {EVA-CLIP-E};
\node[language, right=0.2cm of CogVLM2_vision] (CogVLM2_lang) {LLaMA3 (8B)};

\node[vision, below=1.2cm of CogVLM2_vision] (InstructBLIP_vision) {CLIP ViT-G/14};
\node[language, right=0.2cm of InstructBLIP_vision] (InstructBLIP_lang) {Vicuna (7B, 13B)};

\node[vision, below=1.2cm of InstructBLIP_vision] (InternVL2_vision) {InternViT};
\node[language, right=0.2cm of InternVL2_vision] (InternVL2_lang) {InternLM2};

\node[draw, dashed, rounded corners, inner sep=0.2cm, fit=(coco) (docvqa), label=above:{\textbf{\scriptsize Common Datasets}}] (dataset_group) {};

\node[
    draw,
    dashed,
    rounded corners,
    inner sep=0.2cm,
    minimum width=5cm, 
    fit=(LLaVA_vision) (InternVL2_lang),
    label=above:{\textbf{\scriptsize Vision Encoders and Language Models}}
] (vision_lang_group) {};


\draw[arrow_light] (coco) -- (LLaVA);
\draw[highlight] (scienceqa) -- (LLaVA);

\draw[arrow_light] (coco) -- (InstructBLIP);
\draw[arrow_light] (textcaps) -- (InstructBLIP);
\draw[arrow] (vqav2) -- (InstructBLIP);
\draw[arrow] (okvqa) -- (InstructBLIP);
\draw[arrow] (ocrvqa) -- (InstructBLIP);
\draw[arrow] (visualdialog) -- (InstructBLIP);

\draw[arrow_light] (coco) -- (CogVLM2);
\draw[arrow] (okvqa) -- (CogVLM2);
\draw[arrow] (vqav2) -- (CogVLM2);
\draw[arrow] (docvqa) -- (CogVLM2);
\draw[arrow] (ocrvqa) -- (CogVLM2);
\draw[arrow] (textvqa) -- (CogVLM2);
\draw[highlight] (scienceqa) -- (CogVLM2);
\draw[arrow] (chartqa) -- (CogVLM2);

\draw[arrow_light] (coco) -- (InternVL2);
\draw[arrow_light] (textcaps) -- (InternVL2);
\draw[arrow] (vqav2) -- (InternVL2);
\draw[arrow] (okvqa) -- (InternVL2);
\draw[arrow] (visualdialog) -- (InternVL2);
\draw[highlight] (scienceqa) -- (InternVL2);
\draw[arrow] (docvqa) -- (InternVL2);
\draw[arrow] (ocrvqa) -- (InternVL2);
\draw[arrow] (textvqa) -- (InternVL2);
\draw[arrow] (chartqa) -- (InternVL2);

\draw[arrow] (LLaVA) -- ++(2,0) |- (LLaVA_vision);

\draw[arrow] (CogVLM2) -- ++(2,0) |- (CogVLM2_vision);

\draw[arrow] (InstructBLIP) -- ++(2,0) |- (InstructBLIP_vision);

\draw[arrow] (InternVL2) -- ++(2,0) |- (InternVL2_vision);

\node[legend, above=1.2cm of LLaVA, anchor=north, xshift=2.4cm] (legend) {
\begin{tikzpicture}[inner sep=0, node distance=0.3cm and 0.8cm]
    \node[pretrain, minimum width=0.4cm, minimum height=0.3cm] (p) {};
    \node[right=0.1cm of p, font=\scriptsize] {Training Dataset};
    
    \node[draw=none, right=0.3cm of p] (sep1) {};
    
    \node[model, right=1.5cm of sep1, minimum width=0.4cm, minimum height=0.3cm] (m) {};
    \node[right=0.2cm of m, font=\scriptsize] {Model};
    
    \node[draw=none, right=0.3cm of m] (sep2) {};
    
    \node[vision, right=0.8cm of sep2, minimum width=0.4cm, minimum height=0.3cm] (v) {};
    \node[right=0.1cm of v, font=\scriptsize] {Vision Encoder};
    
    \node[draw=none, right=0.3cm of v] (sep3) {};
    
    \node[language, right=1.5cm of sep3, minimum width=0.4cm, minimum height=0.3cm] (l) {};
    \node[right=0.1cm of l, font=\scriptsize] {Language Model};
    
\end{tikzpicture}
};

\end{tikzpicture}
\caption{Training datasets, vision encoders, and language models for LLaVA, CogVLM2, InstructBLIP, and InternVL2. Non-QA datasets are connected with lighter lines. InternVL2 employs the most diverse QA datasets, enhancing its robustness. Connections to the \emph{ScienceQA} dataset are highlighted. See Appendix for details.}
\label{fig:model_comparison}
\vspace{-0.1in}
\end{figure*}

The performance of the VLMs is influenced by their training datasets and architectural designs. Figure~\ref{fig:model_comparison} summarizes the models' training datasets, vision encoders, and language models. Notably, some models, such as \emph{InternVL2}, are trained on the ScienceQA dataset, raising concerns about potential data contamination. Since the evaluation tasks may overlap with their training data, their performance metrics might be artificially inflated.

The \emph{InternVL2} models combine the InternViT vision encoder with the InternLM2 language model and are trained on a diverse set of datasets, including COCO, VQAv2, OKVQA, Visual Dialog, and ScienceQA. Similarly, \emph{LLaVA} models utilize the CLIP ViT-L/14 vision encoder and \emph{Vicuna} language models, trained on COCO and ScienceQA. In contrast, models like \emph{InstructBLIP} do not include ScienceQA in their training data. They use datasets such as COCO, VQAv2, OKVQA, and Visual Dialog, leveraging the CLIP ViT-G/14 vision encoder and \emph{Vicuna} language models. Their performance is less likely to be influenced by data contamination, providing a more accurate reflection of their capabilities on unseen data.

Overall, while diverse training data and sophisticated architectures contribute to model performance, the inclusion of evaluation datasets in training can artificially inflate results. It is crucial to consider potential data contamination when interpreting performance metrics to ensure fair and accurate assessments of model capabilities.

%% file: sec/7_limitation_and_conclusion.tex
\section{Limitations and Conclusion}
\label{sec:limitations_and_conclusion}

In this paper, we introduced \emph{I-ScienceQA}, a comprehensive benchmark designed to assess the robustness of Vision-Language Models against distractions. By augmenting the \emph{ScienceQA} dataset with diverse forms of distractions, we simulated real-world conditions where input data is often imperfect, noisy. Our extensive evaluation across state-of-the-art VLMs revealed several key findings: (1) Most VLMs remain vulnerable to distractions, especially in the textual domain; (2) Larger models tend to be more robust but do not always guarantee improved performance, particularly when faced with complex bi-modal distractions; (3) Prompt engineering and robust vision encoder could only partially mitigate these vulnerabilities, there remains significant room for improvement. Our findings highlight the need for further research in developing more robust VLM models. As the use of VLMs expands across domains such as healthcare, education, and autonomous systems, it becomes increasingly important to build models that can handle the noisy data often encountered in real-world.

While our work contributes valuable insights into the challenges of distraction robustness, it also has limitations:

\begin{itemize}[leftmargin=2em] \setlength\itemsep{0em} \item Limited scope of distractions: Although we introduced a variety of textual and visual distractions, the dataset does not encompass all possible real-world noise. 

\item Model evaluation focus: Our study primarily focused on pre-trained VLMs. We did not explore the effects of adversary fine-tuning models on noisy datasets that may be more resilient to distractions.

\item Bimodal distractions: While we examined the compounded effects of bimodal distractions, we did not extensively explore how interaction effects between the two modalities influence model performance. 

\item Defense techniques: Although we explored the use of prompt engineering and robust vision encoder as a defense mechanism, our study did not delve into other possible methods to enhance model robustness, such as vision segmentation\cite{lai2023lisa}. 
\end{itemize}

In summary, while the \emph{I-ScienceQA} benchmark provides a valuable tool for evaluating VLM robustness, there is much work to be done in advancing models that can consistently navigate noisy real-world data. Future research should focus on expanding the range of distractions, investigating robust fine-tuning techniques, and exploring defense strategies to improve robustness of VLMs.